\documentclass{article}

\usepackage[final]{corl_2019} 

\usepackage{amsmath}
\usepackage{amsfonts}
\usepackage[inline]{enumitem}
\usepackage{graphicx}
\usepackage{multirow}
\usepackage{booktabs} 

\def\etal{et~al.~}			  
\def\eg{e.g.,~}               
\def\ie{i.e.,~}               

\newcommand*\diff{\mathop{}\!\mathrm{d}}
\DeclareMathOperator{\E}{\mathbb{E}}

\newlength\paramargin
\newlength\figmargin
\newlength\secmargin

\newlength\figwidth

\newcommand{\figref}[1]{Figure~\ref{fig:#1}}

\long\def\ignorethis#1{}

\newcommand{\cutsectionup}{\vspace*{-0.1in}}

\title{Experience-Embedded Visual Foresight}

\author{
  Lin Yen-Chen \qquad Maria Bauza \qquad Phillip Isola \\ 
  Massachusetts Institute of Technology \\
  \texttt{\{yenchenl,bauza,phillipi\}@mit.edu} \\
}

\begin{document}
\maketitle

\begin{abstract}
Visual foresight gives an agent a window into the future, which it can use to anticipate events before they happen and plan strategic behavior. Although impressive results have been achieved on video prediction in constrained settings, these models fail to generalize when confronted with unfamiliar real-world objects. In this paper, we tackle the generalization problem via fast adaptation, where we train a prediction model to quickly adapt to the observed visual dynamics of a novel object.
%
%
%
Our method, Experience-embedded Visual Foresight (EVF), jointly learns a fast adaptation module, which encodes observed trajectories of the new object into a vector embedding, and a visual prediction model, which conditions on this embedding to generate physically plausible predictions. 
For evaluation, we compare our method against baselines on video prediction and benchmark its utility on two real world control tasks.
We show that our method is able to quickly adapt to new visual dynamics and achieves lower error than the baselines when manipulating novel objects. 

%
\end{abstract}



\vspace{-2mm}
\section{Introduction}
\vspace{-2mm}

The ability to visualize possible futures allows an agent to consider the consequences of its actions before making costly decisions. When visual prediction is coupled with a planner, the resulting framework, visual model-predictive control (visual MPC), can perform a multitude of tasks -- such as object relocation and the folding of clothes -- all with the same unified model~\cite{finn2017deep,ebert2018visual}.

Like any learned model, visual MPC works very well when predicting the dynamics of an object it has encountered many times in the past, \eg during training. But when shown a new object, its predictions are often physically implausible and fail to be useful for planning.

This failure contrasts with the ability of humans to quickly adapt to new scenarios. When we encounter a novel object, we may initially have little idea how it will behave -- is it soft or hard, heavy or light? However, just by playing with it a bit, we can quickly estimate its basic physical properties and begin to make accurate predictions about how it will react under our control.

Our goal in this paper is to develop an algorithm that learns to quickly adapt in a similar way. We consider the problem of few-shot visual dynamics adaptation, where a visual prediction model aims to better predict the future given a few videos of previous interactions with an object.

\begin{figure}[t]
    \setlength{\belowcaptionskip}{-4mm}
    \centering
    \includegraphics[width=\linewidth]{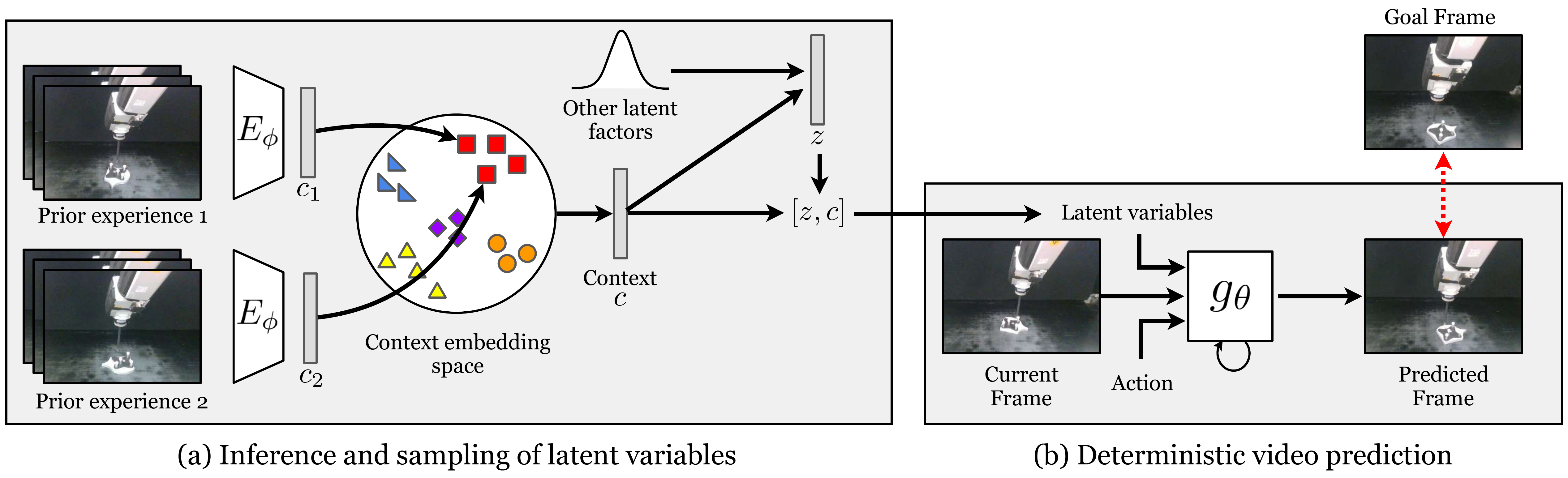}
    \vspace{-4mm}
    \caption{{\bf Experience-embedded Visual Foresight (EVF).} EVF allows a visual prediction model $g_{\theta}$ to better adapt to the visual dynamics of a novel object given a single or multiple videos of this new objects, \ie ``experiences" with it.
    (a) Each experience is first passed through an encoder $E_{\phi}$ to generate a compact representation $c_i$ of the new object, where $i$ is the index of the experience.
    Then, these representations are combined to create a context embedding $c$ which is used to sample latent variable $z$. Both $c$ and $z$ are concatenated along the channel dimension to form an input to the visual prediction model $g_{\theta}$. (b) The visual prediction model is a deterministic recurrent net that predicts the next frame given the current frame, the action taken, the net's recurrent state, and the latent variables inferred and sampled in (a). 
    The experience-embedding encoder $E_{\phi}$ and video prediction model $g_{\theta}$ are jointly optimized.}
    \label{fig:method}
\end{figure}

To that end, we propose Experience-embedded Visual Foresight (EVF), where adaptation is performed via an experience encoder that takes as input a set of previous experiences with the object, in the format of videos, and outputs a low-dimensional vector embedding, which we term a ``context". This embedding is then used as side information for a recurrent frame prediction module. Our method is centered on the idea that there is an embedding space of objects' properties, where similar objects (in terms of visual dynamics) are close together while different objects are far apart. 
Having such a space allows for few-shot visual dynamics adaptation and opens the possibility of inferring information from new and unfamiliar objects in a zero-shot fashion.
Our approach can further be formalized as a hierarchical Bayes model in which object properties are latent variables that can be inferred from a few prior experiences with the object. The full system, as shown in~\figref{method}, is trained end-to-end and therefore can be understood as a form of meta-learning: under this view, the experience encoder is a fast learner, embodied in a feedforward network that modulates predictions based on just a few example videos. The weights of this fast learner are themselves also learned, but by the slower process of stochastic gradient descent over many meta-batches of examples and predictions.

EVF can be categorized as ``learning from observations"~\cite{DBLP:journals/corr/abs-1907-03146,pathak2018zero}, rather than ``learning from demonstrations, as it does not require that observed trajectories contain \emph{actions}. This means it can utilize experiences when actions are unknown or generated by an agent with a different action-space.
To evaluate our approach, we compare EVF against baselines which apply gradient-based meta-learning~\cite{finn2017model} to video prediction, and validate that our method improves visual MPC's performance on two real-world control tasks in which a robot manipulates novel objects. 
%

%
\textbf{Our contributions include}:
\begin{enumerate}
    \item A hierarchical Bayes model for few-shot visual dynamics adaptation.
    \item Improved performance on both video prediction and model-based control.
    \item A learned video representation that captures properties related to object dynamics.
\end{enumerate}

\vspace{-2mm}
\section{Related Work}
\vspace{-2mm}
\paragraph{Embedding-based Meta Learning.}
Learning an embedding function to summarize information from a few samples has shown promising results for few-shot learning tasks.
For image classification, Matching Networks~\cite{vinyals2016matching} learns an embedding space to compare a few labeled examples with testing data, and builds a classifier based on weighted nearest neighbor. Similarly, Prototypical Networks learns an embedding space but instead represents each class by the mean of its examples (a prototype). Then, a classifier is built using squared Euclidean distances between testing data's embedding and different classes' prototypes.
For visuomotor control, TecNets~\cite{james2018task} learns an embedding space of demo trajectories. During test time, it feeds the trajectory embeddings into a control network to perform one-shot imitation learning. We adopt a similar approach but apply it to visual MPC rather than to imitation learning.
For generative modelling, Neural Statistician~\cite{edwards2016towards} extends the variational autoencoder (VAE)~\cite{kingma2013auto} to model datasets instead of individual datapoints, where the learned embedding for each dataset is used for the decoder to reconstruct samples from that dataset.
Our method extends Neural Statistician~\cite{edwards2016towards} to model visual dynamics of diverse objects. The learned model is also used for model-based control.

%
\paragraph{Model-based Reinforcement Learning.}
Algorithms which learn a dynamics model to predict the future and use it to improve decision making belong to the category of model-based reinforcement learning (RL).
Lately, visual model-based RL has been applied to real world robotic manipulation problems~\cite{finn2017deep,ebert2018visual,byravan2017se3,zhang2018solar}.
However, these methods do not focus on adaptation of the visual prediction model.
A few recent works have explored combining meta-learning with model-based RL in order to adapt to different environments in low-dimensional state space~\cite{nagabandi2018learning,nagabandi2018deep}.
In this work, we tackle the problem of adapting high-dimensional visual dynamics by learning a latent space for object properties, inspired by previous works on metric-based meta learning~\cite{vinyals2016matching,james2018task}.

\paragraph{Video Prediction.}
The breakthrough of deep generative models~\cite{goodfellow2014generative,kingma2013auto,oord2016pixel} has lead to impressive results on deterministic video prediction~\cite{mathieu2015deep,villegas2017learning,wichers2018hierarchical,kalchbrenner2017video,finn2016unsupervised}.
Specifically, action-conditioned video prediction has been shown to be useful in video games~\cite{oh2015action,ha2018world} and robotic manipulation~\cite{finn2017deep,ebert2018visual,ebert2017self,xie2018few}. 
Our work is most similar to approaches which perform deterministic next frame generation based on stochastic latent variables~\cite{babaeizadeh2017stochastic,denton2018stochastic,lee2018stochastic,kumar2019videoflow}.
The main difference between these works and ours is shown in~\figref{graphical_models}.
Rather than learning a generative model for all the videos in a single dataset, we construct a hierarchical Bayes model to learn from many small yet related datasets~\cite{edwards2016towards,hewitt2018variational}.
Our goal is to robustify visual MPC's performance by adapting to unseen objects.

\vspace{-2mm}
\section{Preliminaries}
\vspace{-3mm}
In visual MPC (or visual foresight), the robot first collects a dataset $D = \{\tau^{(1)}, \tau^{(2)}, ..., \tau^{(N)}\}$ which contains $N$ trajectories.
Each trajectory $\tau$ can further be represented as a sequence of images and actions $\tau = \{I_1, a_1, I_2, a_2, ..., I_T\}$.
Then, the dataset is used to learn an action-conditioned video prediction model $g_{\theta}$ that approximates the stochastic visual dynamics $p(I_{t}|I_{1:t-1}, a_{t-1})$.
Once the model is learned, it can be coupled with a planner to find actions that minimize an objective specified by user.
In this work, we adopt the cross entropy method (CEM) as our planner and assume the objective is specified as a goal image. 
At the planning stage, CEM aims to find a sequence of actions that minimizes the difference between the predicted future frames and the goal image.
Then, the robot executes the first action in the planned action sequence and performs replanning.
\vspace{-2mm}
\section{Method}
\begin{figure}[t]
    \centering
    \includegraphics[width=\linewidth]{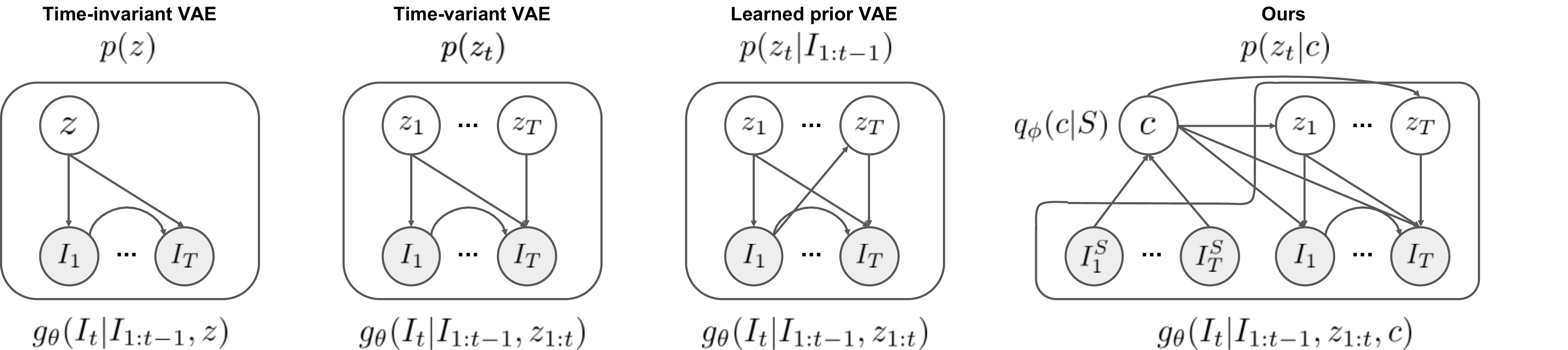}
    \vspace{-4mm}
    \caption{{\bf Graphical models for video prediction.} Actions are omitted for ease of notation. \textit{Left}: time-invariant model proposed in Babaeizadeh~\etal \cite{babaeizadeh2017stochastic}.
    \textit{Middle Left}: time variant model adopted in previous works~\cite{babaeizadeh2017stochastic,denton2018stochastic,lee2018stochastic}.
    \textit{Middle Right}: learned prior model proposed in Denton~\etal\cite{denton2018stochastic}.
    \textit{Right}: our hierarchical generative process for video prediction. Different from previous works~\cite{babaeizadeh2017stochastic,denton2018stochastic,lee2018stochastic}, we introduce a latent variable $c$ that varies between videos of different datasets but is constant for videos of the same dataset. 
    We refer to it as the \textit{context} which captures environmental information used to generate the dataset.
    }
    \label{fig:graphical_models}
\end{figure}

\vspace{-3mm}
Our goal is to robustify visual MPC's performance when manipulating unseen objects through rapid adaptation of visual dynamics.
Inspired by previous works, we hypothesize that humans' capability of rapid adaptation is powered by their ability to infer latent properties that can facilitate prediction.
To endow our agent the same capability, we consider a  setting different from prior work where we are given $K$ datasets $D_k$ for $k \in [1,K]$.
Each dataset $D_k = \{ \tau^{(1)}, \tau^{(2)}, ..., \tau^{(N)} \}$ contains $N$ trajectories of the robot interacting with a specific environment.
We assume there is a common underlying generative process $p$ such that the stochastic visual dynamics $p_k(I_t|I_{1:t-1}, a_{t-1})$ for dataset $D_k$ can be written as $p_k(\cdot) = p(\cdot|c_k)$ for $c_k$ drawn from some distribution $p(c)$, where $c$ is a ``context" embedding.
Intuitively, the context $c$ shared by all trajectories in a dataset $D_i$ captures environmental information that generates the dataset, \eg an object's mass and shape.
Our task can thus be split into generation and inference components. 
The generation component consists of learning a model $g_{\theta}$ that approximates the underlying generative process $p$.
To model stochasticity, we introduce latent variables $z_{t}$ at each time step to carry stochastic information about the next frame in a video.
The inference component learns an approximate posterior over the context $c$. Since encoding the whole dataset $D_i$ is computationally prohibitive, we construct a support set $S$ which contains $M$ trajectories sub-sampled from $D_i$ and learn a function $E_{\phi}(c|S)$ to estimate $c$ from $S$.
In our experiments, $M$ is a small number, $M\leq 5$.
We denote a trajectory, image, and action in the support set as $\tau^S$, $I^S_t$ and $a^S_t$. 
The rightmost subfigure of \figref{graphical_models} shows the graphical model of our method.
Overall, our learning approach is an extension of VAE, and we detail this formalism in Section~\ref{method:vae}.
\vspace{-2mm}
\subsection{Overview}
\vspace{-2mm}
Before detailing the training procedure, we explain how our method performs adaptation and generates predictions for unseen objects. 
To adapt the video prediction model, we first use the inference component, \ie an experience encoder $E_{\phi}(c|S)$ with parameters $\phi$, to encode videos in the support set $S$ into an approximate posterior over context. 
Then, we sample a context from this posterior and feed it into our generation component to produce predictions.
Our generation component is an action-conditioned video prediction model $g_{\theta}$ parameterized by $\theta$, that generates the next frame $\hat{I}_{t}$, based on previous frames $I_{1:t-1}$, action $a_t$, stochastic latent variables $z_{1:t}$, and context $c$.
The two components are trained end-to-end jointly with the aid of a separate frame encoder $q_{\psi}(z_{t}|I_{1:t}, c)$ (not used at test time).
To train our model, we again first sample $c$ from the approximate posterior constructed by the experience encoder. 
Then, the frame encoder encodes $I_{t}$, \ie the target of the prediction model, along with previous frames $I_{1:t-1}$ and $c$ to compute another distribution $q_{\psi}(z_{t}|I_{1:t}, c)$ from which we sample $z_{t}$.
By assuming the prior of the context $c$ and the latent variable $z$ to be normally distributed, we force $E_{\phi}(c | S)$ and $q_{\psi}(z_{t} | I_{1:{t}}, c)$ to be close to the corresponding prior distribution $p(c)$ and $p(z)$ using KL-divergence terms.
These two terms constrain the information that $c$ and $z_{t}$ can carry, forcing them to capture useful information for prediction. 
In addition, we have another term in our loss to penalize the reconstruction error between $\hat{I}_{t}$ and $I_{t}$. 
\subsection{Variational Auto-encoders Over Sets}\label{method:vae}
\vspace{-2mm}
To further explain our model, we adopt the formalism of variational auto-encoders over sets~\cite{edwards2016towards,hewitt2018variational}.
For each trajecory $\tau$, the likelihood given a context $c$ can be written as:
\vspace{-2mm}
\begin{align} 
p(\tau) &= \prod_{t=1}^{T} \int g_{\theta}(I_t | I_{1:t-1}, a_{t-1}, z_{1:t}, c)p(z_{1:t})\diff z_{1:t}
\end{align} 
However, this likelihood cannot be directly optimized as it involves marginalizing over the latent variables $z_{1:t}$, which is generally intractable.
Therefore, the frame encoder $q_{\psi}(z_{t} | I_{1:{t}}, c)$ is used to approximate the posterior with a conditional Gaussian distribution $\mathcal{N}(\mu_{\psi}(I_{1:t}, c), \sigma_{\psi}(I_{1:t}, c))$.
Then, the likelihood of each trajectory in a dataset can be approximated by the variational lower bound:
\begin{align}\label{previous_loss}
L_{\theta, \psi}(I_{1:T}) = \sum_{t=1}^{T} [\E_{q_{\psi}{(z_{1:t}|I_{1:t}, c)}} \log g_{\theta}(I_t | I_{1:t-1}, a_{t-1}, z_{1:t}, c) - \beta D_{KL} (q_{\psi}(z_t|I_{1:t}, c) || p(z))]
\end{align}
Note that although Eq.~\eqref{previous_loss} can be used to maximize the likelihood of trajectories in a single dataset~\cite{babaeizadeh2017stochastic,denton2018stochastic,lee2018stochastic}, our goal is to approximate the commonly shared generative process $p$ through maximizing the likelihood of many datasets.
Based on the graphical model plotted on the right in~\figref{graphical_models}, the likelihood of a single dataset can be written as:
\begin{align} 
p(D) &= \int p(c) \Big[\prod_{\tau \in D} \prod_{t=1}^{T} \int g_{\theta}(I_t | I_{1:t-1}, a_{t-1}, z_{1:t}, c)p(z_{1:t}| c)\diff z_{1:t} \Big] \diff c
\end{align}
In order to solve the intractability of marginalizing over the context $c$, we introduce an experience encoder $E_{\phi}(c | D)$ which takes as input the whole dataset to construct an approximate posterior over context.
By applying variational inference, we can lower bound the likelihood of a dataset with:
\begin{equation}\label{objective}
\begin{split}
L_{\theta, \phi}(D) &= \E_{E_{\phi}(c|D)} \Big[ \sum_{\tau \in D} \sum_{t=1}^{T} [R_D - \beta Z_D] \Big] - \gamma C_D
\ \ \text{where}
\\
R_D &= \E_{q_{\psi}{(z_{1:t}|I_{1:t}, c)}} \log g_{\theta}(I_t | I_{1:t-1}, a_{t-1}, z_{1:t}, c) \\
Z_D &= D_{KL} (q_{\psi}(z_t|I_{1:t}, c) || p(z))
\\
C_D &= D_{KL} (E_{\phi}(c|D) || p(c))
\end{split}
\end{equation}
The hyper-parameters $\beta$ and $\gamma$ represent the trade-off between minimizing frame prediction error and fitting the prior.
Taking smaller $\beta$ and $\gamma$ increases the representative power of the frame encoder and the experience encoder.
Nonetheless, both may learn to simply copy the target frame $I_t$ if $\beta$ or $\gamma$ are too small, resulting in low training error but struggling to generalize at test time.
\subsection{Stochastic Lower Bound}
\vspace{-2mm}
Note that in Eq.~\eqref{objective}, calculating the variational lower bound for each dataset requires passing the whole dataset through both inference and generation components for each gradient update.
This can become computationally prohibitive when a dataset contains hundreds of samples.
In practice, since we only pass a support set $S$ subsampled from $D$, the loss is re-scaled as follows:
\begin{equation}
\begin{split}
\log{p(D)} &\geq \E_{E_{\phi}(c|S)}\Big[ \sum_{\tau \in D} \sum_{t=1}^{T} [R_D - \beta Z_D] \Big] - \gamma C_D =  \sum_{\tau \in D} \Big[ \E_{E_{\phi}(c|S)} \sum_{t=1}^{T} [R_D - \beta Z_D] - \frac{1}{|D|} \gamma C_D \Big]
\end{split}
\end{equation}
In this work, we sample 5 trajectories without replacement to construct the support set $S$.

\subsection{Model Architectures}
\vspace{-2mm}
We use SAVP~\cite{lee2018stochastic} as our backbone architecture for the generation component.
Our model $g_{\theta}$ is a convolutional LSTM  which predicts the transformations of pixels between the current and next frame.
Skip connections with the first frame are added as done in SNA~\cite{ebert2017self}. 
Note that at every time step $t$, the generation component only receives $I_{t-1}$, $a_{t-1}$, $z_t$, and $c$ as input.
The dependencies on all previous $I_{1:t-2}$ and $z_{1:t-1}$ stem from the recurrent nature of the model.
To implement the conditioning on action, latent variables, and context, we concatenate them along the channel dimension to the inputs of all the convolutional layers of the LSTM.
The experience encoder is a feed-forward convolutional network which encodes the image at every time step and an LSTM which takes as input the encoded images sequentially.
We apply the experience encoder to $M$ trajectories in the support set to collect $M$ LSTM final states.
Sample mean is used as a pooling operation to collapse these states into the mean of our approximate posterior $E_{\phi}(c|S)$.
The frame encoder is a feed-forward convolutional network which encodes $I_t$ and $I_{t-1}$ at every time step. 
We train the model using the re-parameterization trick and estimate the expectation over $q_{\psi}(z_{1:t} | I_{1:t}, c)$ with a single sample.
\section{Experiments}
\vspace{-3mm}
In this section, we aim to answer the following questions: 
\begin{enumerate*}[label=(\roman*)]
  \item Can we learn a context embedding $c$ that facilitates visual dynamics adaptation?
  \item Does the context embedding $c$ improve the performance of downstream visual MPC tasks?
  \item Can the context embedding $c$ be used to infer similarity between objects?
\end{enumerate*}
First we present video prediction results on Omnipush and KTH Action to answer questions (i) and (iii) from section~\ref{exp:omnipush} to~\ref{exp:kth}. 
Then, we report results on two real world control tasks to answer question (ii) in section~\ref{exp:control}.
In our experiments, we compare our method to the following baselines:
\begin{itemize}[wide, labelwidth=!, labelindent=0pt]
\item \textbf{SAVP. } Stochastic video prediction model trained with the VAE objective proposed in~\cite{lee2018stochastic}. We only use VAE loss as it's more intuitive to apply finetuning or gradient-based meta learning~\cite{finn2017model} to it. Also, we found no notable performance difference.
\item \textbf{SAVP-F. } SAVP finetuned with 5 videos in the support set with learning rate 5e-4 for 50 steps.
\item \textbf{SAVP-MAML. } SAVP trained with MAML~\cite{finn2017model} and finetuned the same way as in SAVP-F.
We also tried implementing an RNN-based meta-learner~\cite{duan2016rl}, where SAVP's hidden states are not reset after encoding videos from the support set. However, we were not able to achieve improved performance using this method.
\end{itemize}
\subsection{Omnipush Dataset}\label{exp:omnipush}
\vspace{-3mm}
Our first experiment use Omnipush dataset~\cite{omnipush} which contains RGBD videos of an ABB IRB 120 industrial arm pushing 250 objects, each pushed 250 times.
For each push, the robot selects a random direction and makes a 5cm straight push for 1s.
Using modular meta-learning~\cite{alet2018modular} on ground truth pose data (rather than raw RGBD), the authors showed that Omnipush is suitable for evaluating adaptation because it provides systematic variability, making it easy to study, \eg the effects of mass distribution vs. shape on the dynamics of pushing.  
%

\begin{figure}[]
    \centering
    \includegraphics[width=\linewidth]{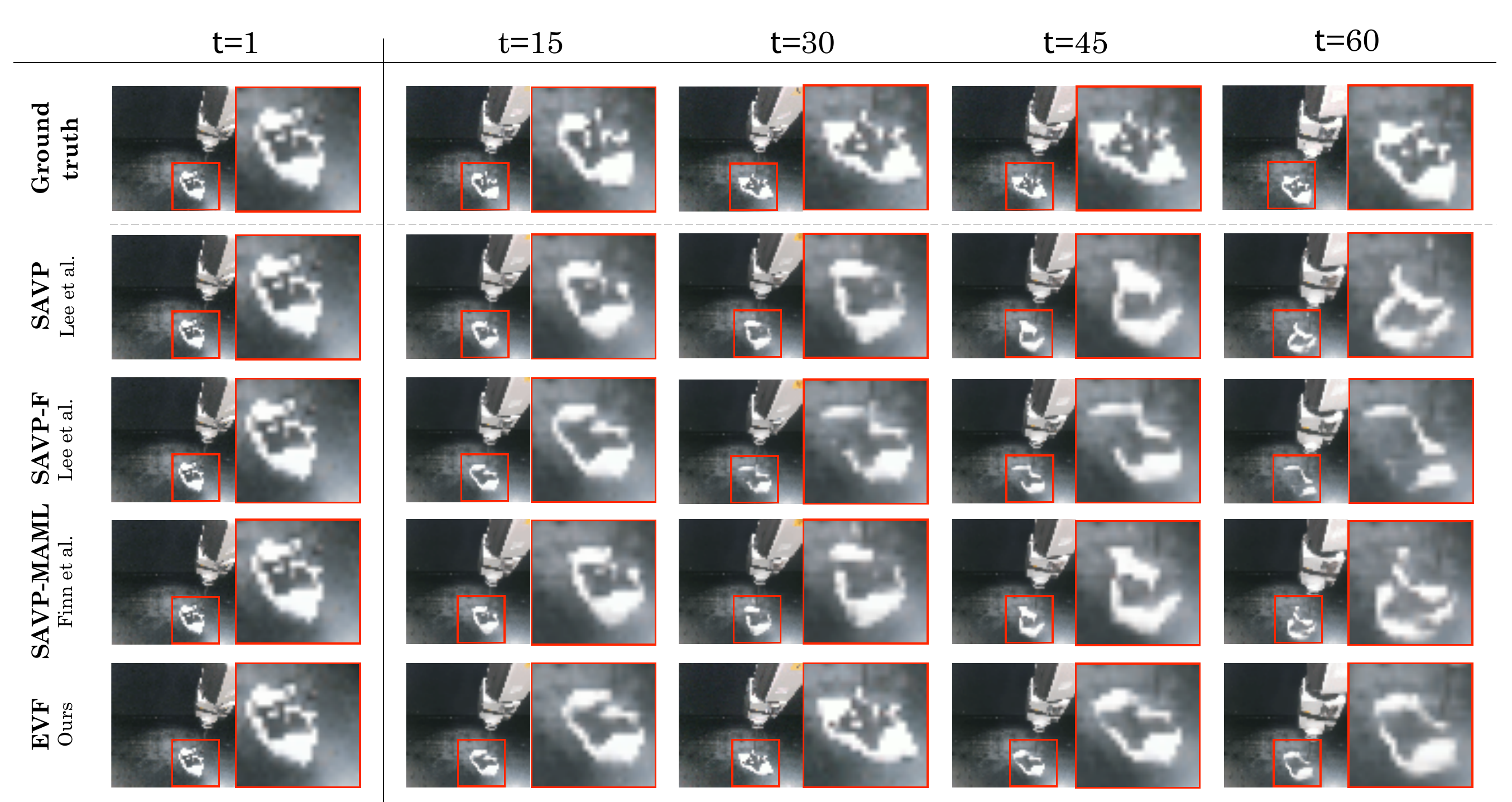}
    \vspace{-6mm}
    \caption{{\bf Qualitative Results for Omnipush.}
    We show generated videos from EVF (ours), SAVP, SAVP-F, and SAVP-MAML.
    For clarity, patches that contain objects are cropped and shown on the right of predicted frames at each time step.
    Each method predicts 60 frames, with the time steps indicated at the top. 
    \textit{Left}: SAVP predictions deform the novel object into a circle.
    For both SAVP-F and SAVP-MAML, the novel object vanishes into the background.
    In comparison, EVF is able to predict more physically plausible futures.
    \textit{Right}: the novel object completely deforms into an ellipse in all the predictions generated by baseline methods.
    Comparatively, EVF preserves the orientation and the shape of object better.
    This can be useful for visual MPC whose robustness heavily relies on the correctness of the predictions.
    Please refer to our supplementary material for video results.
    }
    \label{fig:qualitative_omnipush}
\end{figure}

\begin{figure}[t]
    \setlength{\belowcaptionskip}{-4mm}
    \centering
    \includegraphics[width=\linewidth]{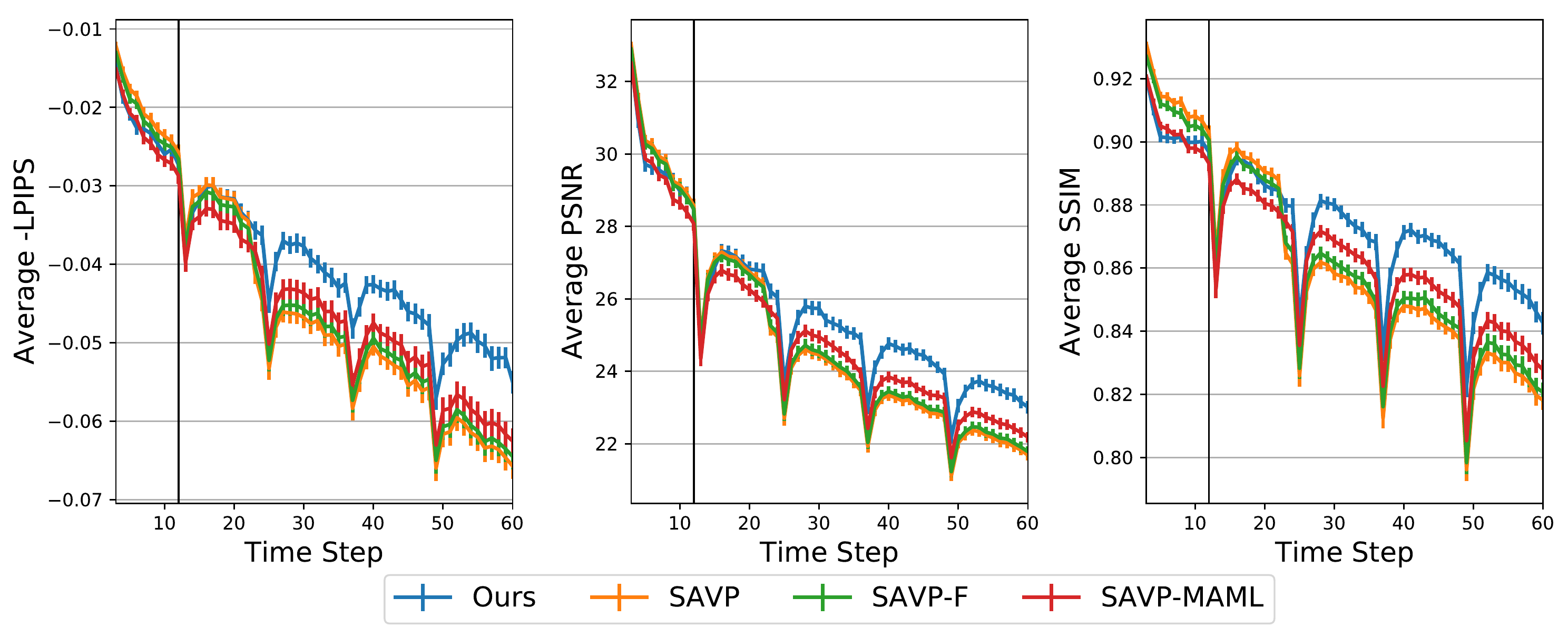}
    \vspace{-6mm}
    \caption{{\bf Quantitative Results for Omnipush.} We show the similarity (higher is better) between ground truth and the best sample as a function of prediction time step. Three metrics -- LPIPS (\textit{left}), PSNR(\textit{middle}), and SSIM (\textit{right}) are used in our evaluation. Spikes appeared every 12 time steps are caused by the switch of pushing direction in Omnipush dataset.}
    \label{fig:quantitative_omnipush}
\end{figure}

\begin{figure}[t!]
    \setlength{\belowcaptionskip}{-4mm}
    \centering
    \includegraphics[width=0.8\linewidth]{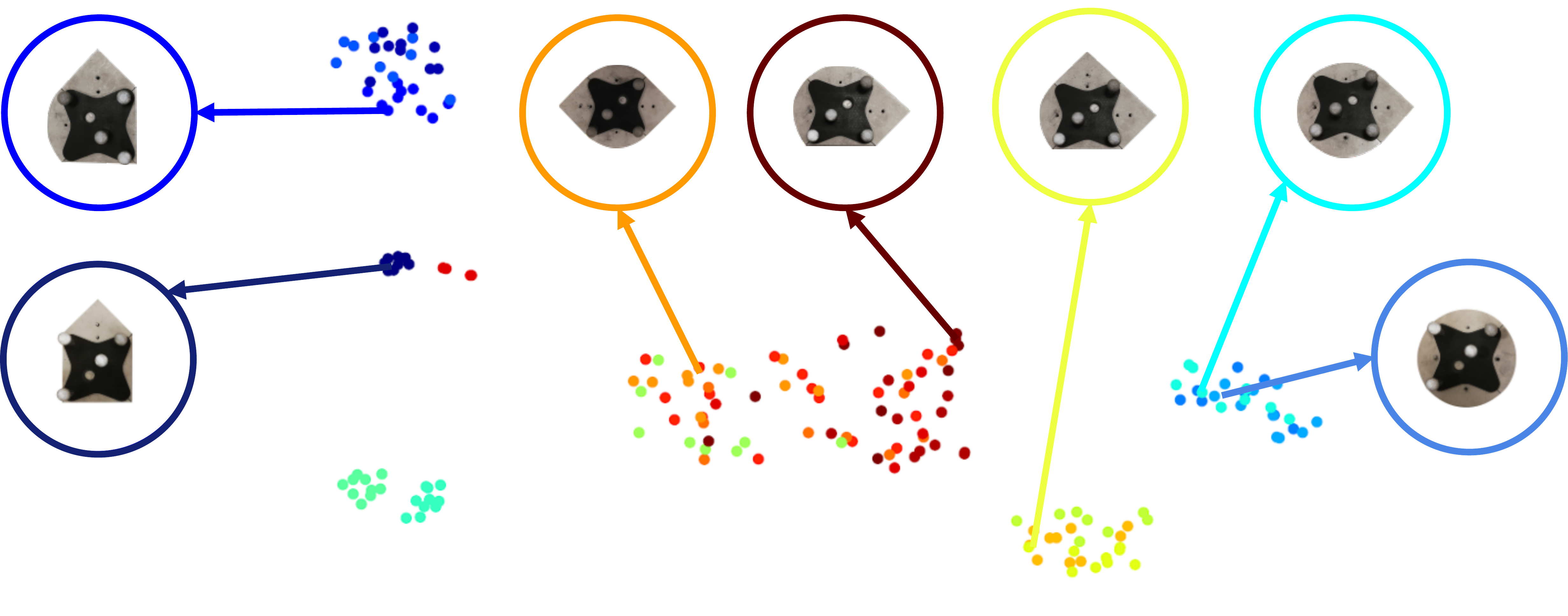}
    \caption{{\bf t-SNE~\cite{maaten2008visualizing} results for Omnipush.} We visualize the context embedding of 20 novel objects through t-SNE. We found that embeddings are closer to each other when objects posses similar shapes and mass.}
    \label{fig:tsne_omnipush}
\end{figure}

%
We use the first split of the dataset (\ie 70 objects without extra weight) to conduct the experiments.
We condition on the first 2 frames and train to predict the next 10.
To benchmark the adaptability of our approach, we train on 50 objects and test on the remaining 20 novel objects.
\figref{qualitative_omnipush} shows qualitative results of video prediction.
Following previous works~\cite{denton2018stochastic,lee2018stochastic}, we provide quantitative comparisons by computing the learned perceptual image patch similarity (LPIPS)~\cite{zhang2018unreasonable}, the structural similarity (SSIM)~\cite{wang2004image}, and the peak signal-to-noise ratio (PSNR) between ground truth and generated video sequences.
Quantitative results are shown in~\figref{quantitative_omnipush}.
Our method predicts more physically plausible videos compared to baselines, and especially captures the shape of the pushed object better. We further show that this improves the performance of visual MPC in section~\ref{exp:control}.
\paragraph{Visualization of context embedding.}
In~\figref{tsne_omnipush}, we use t-SNE~\cite{maaten2008visualizing} to visualize the context embedding of the 20 novel objects we tested.
Despite never seeing these object during training, our model is able to produce a discriminative embedding $c$ where objects with similar properties, such as mass and shape, are close together, and those with different properties are far apart. 
%
\begin{figure}[]
    \setlength{\belowcaptionskip}{-4mm}
    \centering
    \includegraphics[width=\linewidth]{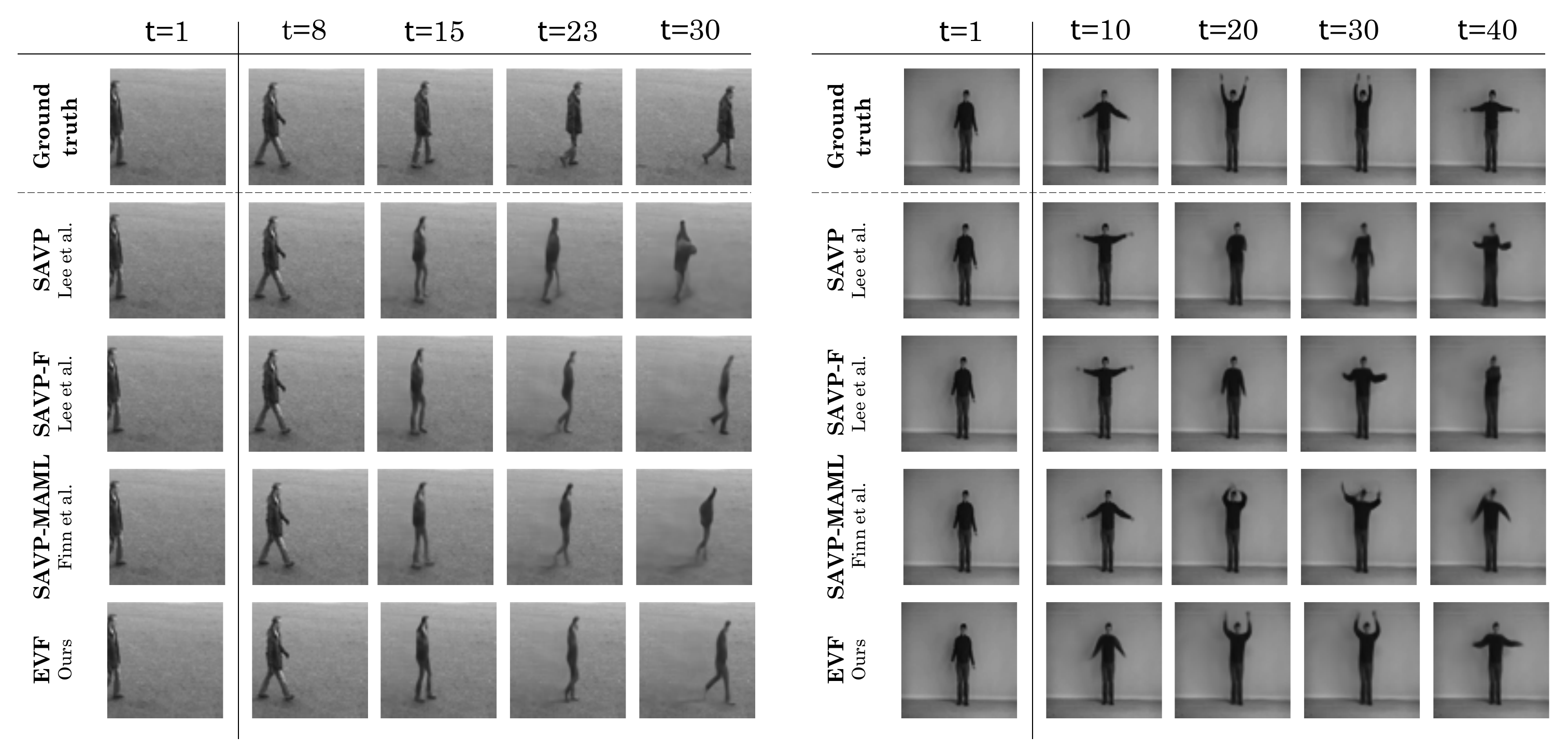}
    \vspace{-6mm}
    \caption{{\bf Qualitative Results for KTH Action.}
    We show generated videos from EVF (ours), SAVP, SAVP-F, and SAVP-MAML.
    Each method predicts 30 frames, with the time steps indicated at the top. 
    \textit{Left}: SAVP's predictions deform the human body.
    For SAVP-MAML and SAVP-F, the human body and head are compressed in the latter time steps.
    In comparison, EVF is able to predict more realistic futures.
    \textit{Right}: For both SAVP and SAVP-F, human's hand vanishes into its body.
    SAVP-MAML's predictions preserve human's hand but slightly deform its head. 
    In comparison, EVF preserves the shape of the human better.
    Results on KTH show that our method can be applied to domains beyond robotics.
    }
    \label{fig:qualitative_kth}
\end{figure}
\begin{figure}[t]
    \centering
    \includegraphics[width=\linewidth]{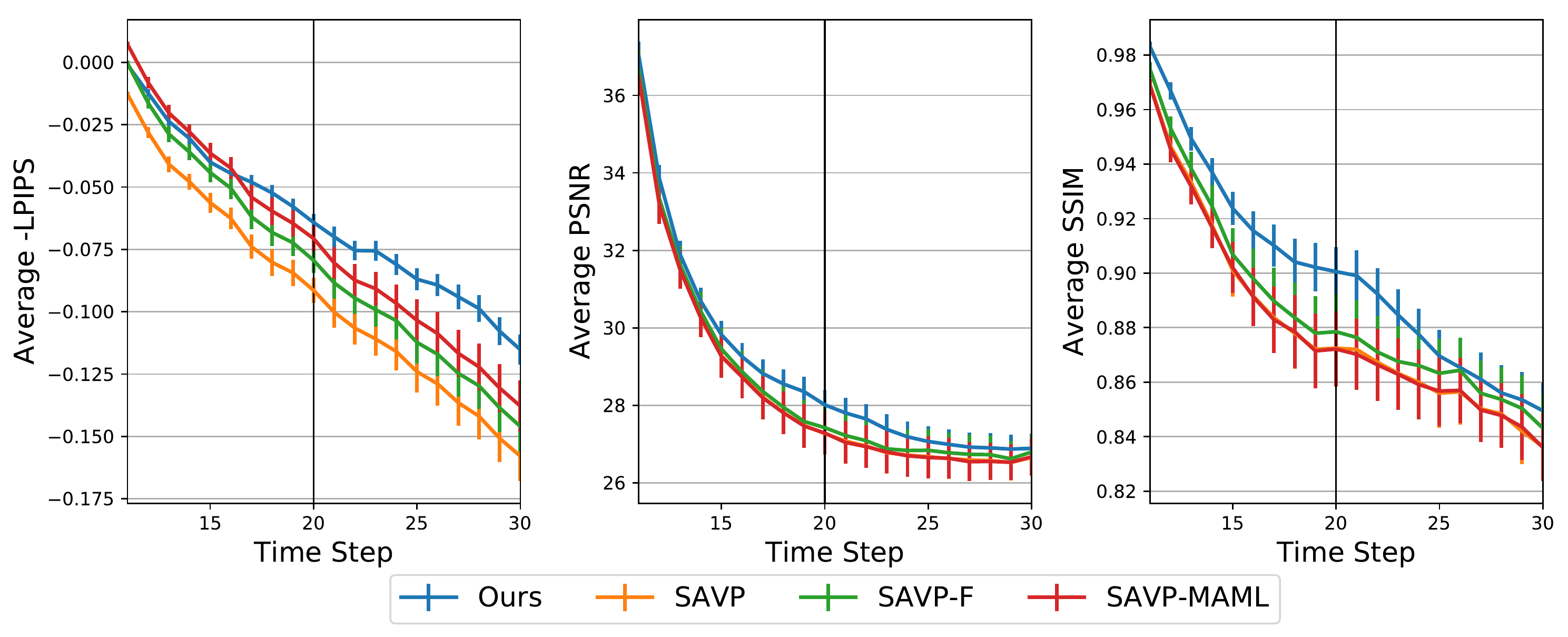}
    \vspace{-4mm}
    \caption{{\bf Quantitative Results for KTH Action.} We show the similarity (higher is better) between ground truth and the best sample as a function of prediction time step. Three metrics -- LPIPS (\textit{left}), PSNR(\textit{middle}), and SSIM (\textit{right}) are used in our evaluation. .}
    \label{fig:quantitative_kth}
\end{figure}
\cutsectionup
\subsection{KTH Action Dataset}\label{exp:kth}
\vspace{-2mm}
The KTH Action dataset~\cite{schuldt:04} consists
of real-world videos of people performing different actions, where each person performs the same action in 4 different settings.
We split the dataset into many small datasets according to person and action.
Therefore, the support set consists of the other 3 videos belonging to the same small dataset.
Following the setting of previous work~\cite{lee2018stochastic}, our models are trained to predict 10 future frames conditioned on 10 initial frames.
During test time, it is tested to predict 30 frames into the future.
We show qualitative results in ~\figref{qualitative_kth} and quantitative results in~\figref{quantitative_kth}.
\subsection{Real World Pushing}\label{exp:control}
\vspace{-2mm}
We perform two types of real-world robotic experiments to verify that predictions generated by our method are more physically plausible, and that this benefits visual MPC.
In the first experiment, we ask the robot to perform re-location, where the robot should push an object to a designated pose.
In the second experiment, the robot has to imitate the pushing trajectory presented in a 10-second video.
In both tasks we use pixel-wise $\ell_2$ loss, with the robot arm masked, to measure the distance between the predicted image and the goal image.
At each iteration of planning, our CEM planner samples 200 candidates and refits to the best 10 samples.
The results of this experiment can be found in Table~\ref{table:results}.
Note that several more advanced loss functions have been recently proposed for visual MPC~\cite{xie2018few,ebert2017self,ebert2018visual}, but we opted for simple $\ell_2$ to isolate the contribution of our method in terms of improved visual predictions, rather than improved planning.
\begin{table}[t]
  \centering
  \tabcolsep=0.1cm{
  \caption{\textbf{Errors in re-positioning and trajectory-tracking task.} Units are millimeter. \textit{Left}: in re-positioning, results are calculated with 5 different random seeds. \textit{Right}: in trajectory-tracking, we report the mean error across time steps and the final offset.}\label{table:results}
  \begin{tabular}{llll}
    \multicolumn{4}{c}{\textbf{Re-positioning}} \\
    \toprule[1pt]
      Method & Type & Seen  & Unseen\\ 
    \midrule[1pt]
      \multirow{ 2}{*}{No motion }
      & mean & 90.4  & 91.9 \\
      & median & 87.4  & 90.4 \\
      \midrule[0.3pt]
      \multirow{ 2}{*}{SAVP } 
      & mean & 26.2  & 48.4 \\
      & median & 27.1  & 48.5 \\ 
      \midrule[0.3pt]
      \multirow{ 2}{*}{EVF (ours)}
      & mean  & \textbf{21.8}  & \textbf{36.2} \\ 
      & median & \textbf{20.1} & \textbf{29.1} \\ 
    \bottomrule[1pt]
  \end{tabular}
  \quad
  \begin{tabular}{llllll}
    \multicolumn{6}{c}{\textbf{Trajectory-tracking}} \\
    \toprule[1pt]
      Method & Type & Traj 1  & Traj 2 & Traj 3  & Traj 4 \\ 
    \midrule[1pt]
      \multirow{ 2}{*}{No motion } 
      & mean & 112  & 93.6  & 91.1  & 97.7 \\
      & final &154  & 127  & 137  & 141 \\
      \midrule[0.3pt]
      \multirow{ 2}{*}{SAVP } 
      & mean & 33.5  & 42.7  & 29.5  & 41.2 \\
      & final & 79.8  & 60.8  & 49  & 59.5 \\ 
      \midrule[0.3pt]
      \multirow{ 2}{*}{EVF (ours)}
      & mean  & \textbf{31.6}  & \textbf{25.2}  & \textbf{21.6}  & \textbf{27.5}  \\ 
      & final & \textbf{59.9}  & \textbf{47.7}  & \textbf{48.7}  & \textbf{48.6} \\ 
    \bottomrule[1pt]
  \end{tabular}}
  \newline
  \newline
  \vspace{-0.5\baselineskip}
  \vspace{-4mm}
\end{table}
\vspace{-2mm}
\section{Conclusion}
\vspace{-2mm}
In this work, we considered the problem of visual dynamics adaptation in order to robustify visual MPC's performance when manipulating novel objects.
We proposed a hierarchical Bayes model and showed improved results on both video prediction and model-based control.
In the future, it would be interesting to investigate how to acticely collect useful data for few-shot adaptation.
	


\clearpage
\acknowledgments{We thank Alberto Rodriguez, Shuran Song, and Wei-Chiu Ma for helpful discussions. This research was supported in part by the MIT Quest for Intelligence and by iFlytek.}


\bibliography{example}  

\end{document}